\documentclass{article}
\usepackage{graphicx}
\usepackage[utf8]{inputenc}
\usepackage{hyperref}
\usepackage{listings}
\usepackage{xcolor}
\usepackage{combelow}
\graphicspath{ {./images/} }

\lstdefinelanguage{openscad}{
  keywords={abs, sign, sin,cos,tan,acos,asin,atan,atan2,floor,round,
ceil,ln,len,let,log,pow,sqrt,exp,rands,min,max, union,difference,intersection,sphere,cube,cube,
cylinder,polyhedron,translate,rotate,scale,resize,mirror,multmatrix,color,color,offset,hull,
minkowski,circle,square,polygon,text,
var, module, function, include, use, echo, for, projection, intersection_for, linear_extrude, rotate_extrude, surface, render, children, import, if, else,\$fn, \$fa, \$fs, \$t, 
concat,lookup,str,chr,search,version,version_num,norm,cross,parent_module},
  keywordstyle=\color{blue}\bfseries,
  ndkeywords={},
  ndkeywordstyle=\color{darkgray}\bfseries,
  identifierstyle=\color{black},
  sensitive=false,
  comment=[l]{//},
  morecomment=[s]{/*}{*/},
  commentstyle=\color{purple}\ttfamily,
  stringstyle=\color{red}\ttfamily,
  morestring=[b]',
  morestring=[b]",
  breaklines=true,
}

\begin{document}

\pagestyle{headings}

\title{Jenny 5 - the robot}

\author{Mihai Oltean\\
Faculty of Exact Sciences and Engineering,\\
"1 Decembrie 1918" University of Alba Iulia,\\
Alba-Iulia, Romania.\\
mihai.oltean@gmail.com \\
\url{https://jenny5.org}\\
\url{https://jenny5-robot.github.io}
}

\maketitle

\begin{abstract}

Jenny 5 is a fully open-source robot intended to be used mainly for research but it can act as a human assistant too. It has a mobile platform with rubber tracks, a flexible leg, two arms with 7 degrees of freedom each and head with 2 degrees of freedom. The robot is actuated by 20 motors (DC, steppers and servos) and its state is read with the help of many sensors. The robot also has 3 webcams for computer vision tasks. In this paper the current state of the robot is described.

\end{abstract}
\newpage
\tableofcontents 
\newpage
\section{Introduction}

Jenny 5 represents an attempt to build a low-cost, almost humanoid robot, by using tools and materials which are readily available.

Jenny 5 is inspired by the Johnny 5 robot from the Short circuit movie \cite{short_circuit_movie}.

Work to Jenny 5 robot was started in April 2015. Until now 3 major versions have been built and tested. At each iteration the robustness of the robot has been significantly improved.

\begin{figure} 
  \includegraphics[width=\textwidth]{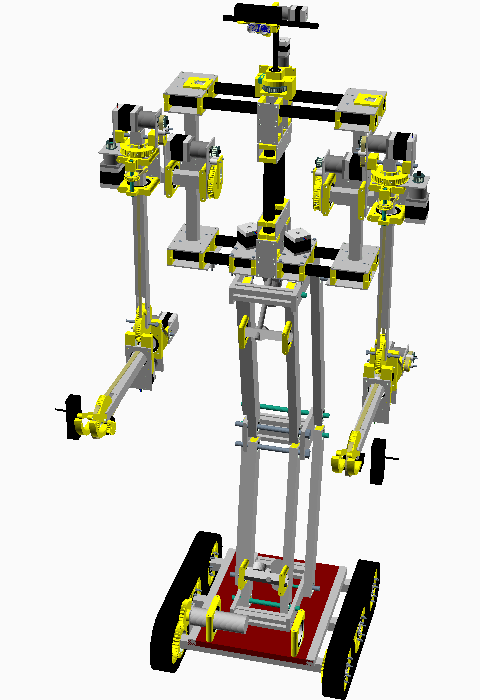}
  \caption{Jenny 5 robot. Design view}
  \label{jenny5_design}
\end{figure}

\begin{figure} 
  \includegraphics[width=\textwidth]{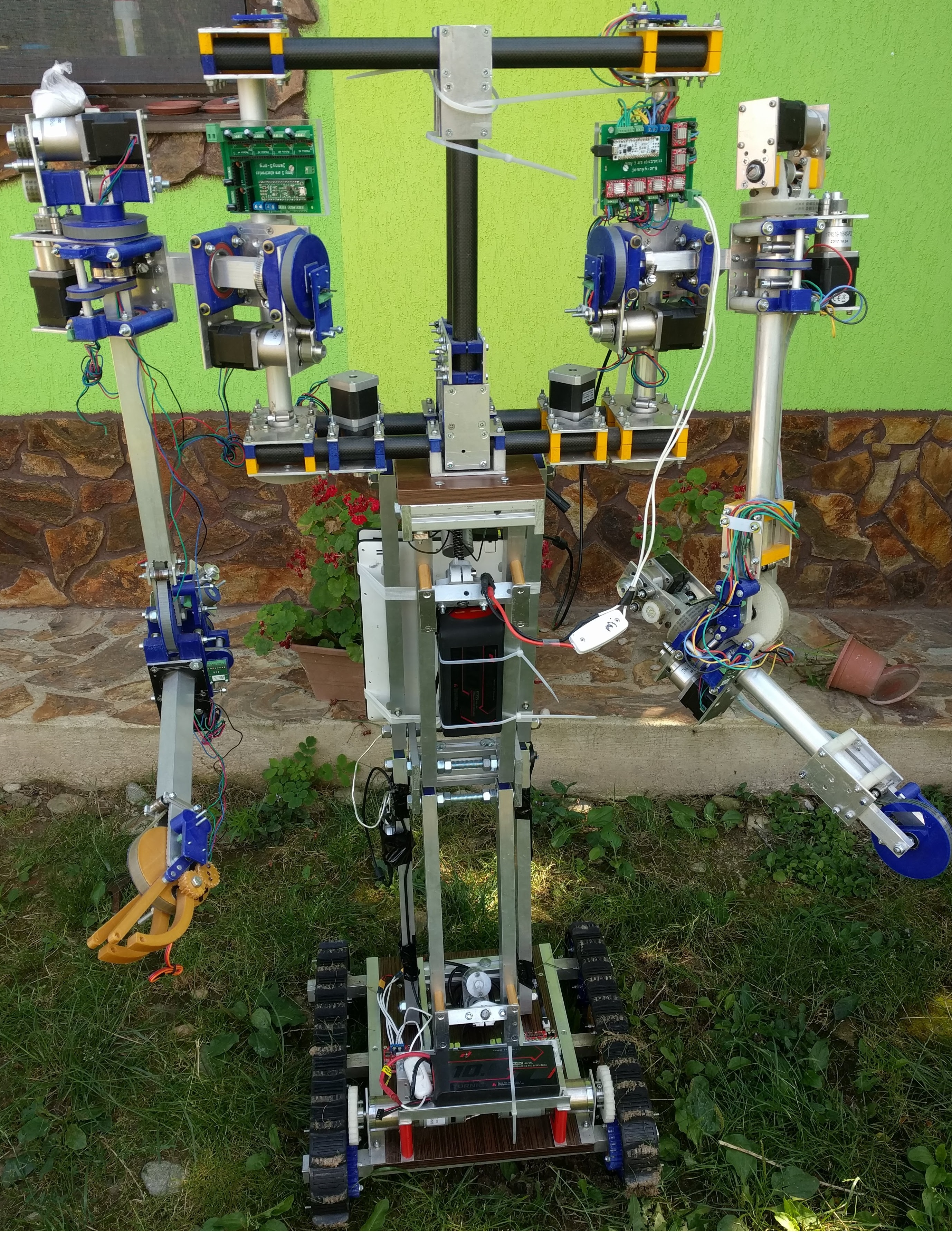}
  \caption{Jenny 5 robot. Real world view of the current iteration (v3).}
  \label{jenny5_real}
\end{figure}

Jenny 5 (v3) has a mobile platform with tracks, a pliable leg, two arms with 7 degrees of freedom each and one head. The current design of the Jenny 5 is given in Figure \ref{jenny5_design}. A real world image is given in Figure \ref{jenny5_real}.

All source files (CAD, software etc) for Jenny 5 are freely available on GitHub \cite{jenny5_source_code,jenny5_website}. All code is released under MIT license so that anyone can freely use it in both personal and commercial applications.

Several programming languages have been used for developing Jenny 5 hardware and software. These are: C++ (for server and the Arduino firmware), HTML5 and JavaScript (for the client controlling the server) and OpenSCAD (for the hardware design).

Jenny 5 is easy and cheap to build. Most components can be purchased from robotics stores. Some aluminum profiles can be cut and drilled with tools available for hobbyists. Custom made parts can be printed with a 3D printer.

Materials (excepting the onboard computer) cost less than 2500 USD.

The robot can be utilized in a wide range of scenarios. Here is a short list of things that the robot could do (if programmed properly): surveillance, rescue, disasters management, house cleaning, food preparation, cleaning kitchen table, working in the garden, fire fighting, combat missions etc.

The Jenny 5 robot is continuously developed and updated. This paper describes the current state of the robot (also called v3). 

The document is structured as follows: 

Section \ref{source_code} contains the locations where the source code of the robot is stored. 

Section \ref{specifications} contains a list of technical specifications for Jenny 5. 

Section \ref{hardware_design}  introduces \textit{OpenSCAD} which is the software used to design the robot. Later the CAD project structure is given. 

Section \ref{hardware} describes the main components of the robot (mobile platform, leg, arms, body and head).  A short list of materials and necessary tools is given in sections \ref{materials} and \ref{tools}. 

Section \ref{electronics} describes the electronic boards used for reading sensors and for moving motors. 

Section \ref{brain} gives the specifications of the computer controling the robot.

Batteries utilized in this project are listed in section \ref{power_supply}. 

Section \ref{jenny5_software} describes the software that controls the robot (\textit{Scufy} - the Arduino firmware, \textit{Scufy Lib} - Arduino PC control library and the HTML 5 and WebSocket server). Several intelligent algorithms used by Jenny 5 are described in section \ref{Intelligent_algorithms}. 

Section \ref{Future_work} contains a short description of some future development directions.

\newpage
\section{Source code}
\label{source_code}

Jenny 5 is open source and the entire source code is stored as a GitHub organization at: \url{https://github.com/jenny5-robot/}. Individual repositories can be downloaded from the above-indicated address. The Bill of Materials (BOM) and the assembly manual are also available online. 

The source code is released under MIT license \cite{mit_license} so that anyone can use it free or paid, open-source or closed-source, private or public projects. The only request is to mention the author(s).

Below is the list of repositories and their web addresses:

\begin{itemize}
    \item \textit{CAD files} - can be downloaded from \url{https://github.com/jenny5-robot/jenny5-cad}. It requires \textit{OpenSCAD} \cite{openscad_website} to view.

    \item \textit{Scufy} - the Arduino firmware - can be downloaded from \url{https://github.com/jenny5-robot/Scufy}. It is used by the arm and head electronics. It requires Arduino IDE \cite{arduino_website} to run and upload to boards.
    
    \item \textit{Scufy Lib} - Arduino control library - can be downloaded from \url{https://github.com/jenny5-robot/jenny5-control-module}. It is used for sending commands to Arduino firmware from a PC. It is written in C++ and requires a C++ compiler.
    
    \item \textit{WebSocket server and HTML5 client} -  can be downloaded from \url{ https://github.com/jenny5-robot/jenny5-html5}. It is used to control the robot from a browser. The server is a console application running on the robot PC.
    
    \item \textit{custom PCBs} - can be downloaded from \url{https://github.com/jenny5-robot/jenny5-electronics}. They require Fritzing \cite{fritzing} to view or modify.
    
    \item \textit{the assembly manual} - can be read from \url{https://jenny5.org/manual/}. Note that currently this is under construction and will not be completed until the robot is completed.
    
    \item The \textit{Bill of Materials (BOM)} is stored as a Google Spreadsheet at: \url{https://docs.google.com/spreadsheets/d/1SwaC4woJEvUEqn66lIG9MLYkuADfTsWIQtvDKaCQT1s/edit?usp=sharing}.

\end{itemize}

\newpage
\section{Specifications}
\label{specifications}

Technical specifications are given in Table \ref{tab:table_specs}. Please note that some specifications are not computed, but based on my experience with the robot.

\begin{table}[ht]
    \centering
    \begin{tabular}{p{5cm}p{5cm}}
    \hline
         Parameter&Value  \\
    \hline\hline
    Total Weight & 40kg\\
    \hline
    Width & 80cm (from left arm extremity to right arm extremity)\\
    \hline
    Total height & 1.8m (from ground to head tip)\\
    \hline
    Leg height & 35cm (fully compressed) to 95cm (fully extended)\\
    \hline
    Arm length (from shoulder connection to gripper center) & 70 cm\\
    \hline
    Arm weight lift & $<$1kg (when fully extended). Can more if a spring is used. See section \ref{arms_section} for a discussion.\\
    \hline
    Arm (time for a up-down move)&11 seconds. See section \ref{arms_section} for a discussion.\\
    \hline
    Speed & $<$3km/h. Can run faster if a motor with a lower gear reduction is used.\\
    \hline
    Leg (time to compress/extend)&7 seconds.\\
    \hline
    Autonomy&Several hours (depends on the capacity of the batteries).\\
    \hline
    \end{tabular}
    \caption{Technical specifications of the current iteration of Jenny 5. Please note that some specifications can be improved when using better components.}
    \label{tab:table_specs}
\end{table}

\newpage
\section{Hardware design}
\label{hardware_design}

This section describes the software used to design Jenny 5 and the structure of the CAD project.

\subsection{CAD software}
\label{openscad}

Jenny 5 is designed in \textit{OpenSCAD} \cite{openscad_website}. I have chosen this environment because it is mainly intended for programmers and I (the author) am a programmer.

Parts in \textit{OpenSCAD} are designed using computer instructions instead of using the mouse.

\textit{OpenSCAD} IDE is very simple: one writes instructions in the left window, and the result (after pressing \textit{F5}) are seen in the top-right window. In the bottom-right window the programmer can display various information by using \textbf{echo} instruction.

The language offers several primitives (\textbf{cube}, \textbf{cylinder}, \textbf{sphere} etc), several Boolean operations (\textbf{union}, \textbf{difference}, \textbf{intersection}) and several transformations (\textbf{translate}, \textbf{rotate}, \textbf{mirror}, etc). By using only these primitives and operations is possible to create very complex geometries.

The programmer can write everything in the main file, but usually the instructions creating a part are grouped within a \textbf{module}.

For instance a simple program which creates a nut is the following:\newline

\begin{lstlisting}[language = openscad]

module nut(external_radius, internal_radius, thickness)
{
    difference(){
        cylinder(h = thickness, r = external_radius, $fn = 6);
        cylinder(h = thickness, r = internal_radius, $fn = 50);
    }
}
// now call the module
nut(external_radius = 4, internal_radius = 2, thickness = 3.2);
\end{lstlisting}

For visualizing a part the user should press \textit{F5} (\textit{Preview}) or \textit{F6} (\textit{Render}). 

One can use the mouse to navigate through the designed part. Left button drag is used for rotate, right button drag for move and mouse wheel for zoom.

Note that for visualizing the entire robot we will use \textit{Preview} mode only because \textit{Render} is too slow (can take several hours to render the entire robot). However, when we need the \textit{stl} file for 3D printing, we will use \textit{Render} mode for that particular part.

The user can uncomment each module of the project to show it on screen.

\subsection{CAD project structure}

CAD files for the robot can be downloaded from \url{https://github.com/jenny5-robot/jenny5-cad}.

There are 2 main folders there: \textit{basic\_scad} and \textit{robot}. 

\begin{itemize}
    \item \textit{basic\_scad} - contains general components like screws, nuts, motors, sensors, pulleys, gears, bearings, etc. These components can be used in other projects too.
    \item \textit{robot} - contains components specific to Jenny 5 robot. Each part of the robot has its own folder. Main file is called \textit{jenny5.scad}. To view it just open it in \textit{OpenSCAD} and press \textit{F5}.
\end{itemize}

The \textit{robot} folder contains 5 subfolders:

\begin{itemize}
\item \textit{arm}. The main file is \textit{arm.scad}.
\item \textit{base\_platform}. The main file is \textit{base\_platform.scad}
\item \textit{body}. The main file is \textit{body.scad}. Please note that arms are connected to body only in the main file of the project (\textit{jenny5.scad} ).
\item \textit{head}. The main file is \textit{head.scad}.
\item \textit{leg}. The main file is \textit{leg.scad}.
\end{itemize}

\subsubsection{Parameters}

The current position of the leg and arms is stored as numerical values (angles).

These parameters are stored in the following files:

\begin{itemize}
    \item 
\textit{robot/arm/arm\_params.scad} - here we have the angles for arms.

\item
\textit{robot/leg/leg\_params.scad} - here we have the angle for leg.

\end{itemize}

One can modify these parameters in order to simulate a new position for leg or arms.

One can change the color of the plastic parts from file \textit{basic\_scad/material\_colors.scad}.

\newpage
\section{Hardware}
\label{hardware}

This section describes the most important mechanical components of the robot which are the mobile platform (see section \ref{mobile_platform_section}), the leg (see section \ref{leg_section}), two arms (see section \ref{arms_section}), a body (see section \ref{body_section}) and a head (see section \ref{head_section}).

\subsection{Materials}
\label{materials}

In Table \ref{tab:materials} the materials needed for the mechanical parts are listed. Please note that electrical / electronic components (including sensors, cameras, micro-controllers, etc) are not listed here, but the motors are. The quantities are not always exact. For more details the user is encouraged to read the online assembly manual from \cite{jenny5_website}.

\begin{table}[ht]
    \centering
    \begin{tabular}{p{5cm}p{5cm}}
    \hline
         Material&Quantity  \\
    \hline\hline
    M3, M4, M8, M12 screws&Many (of various lengths)\\ 
    \hline
    M3, M4, M6, M8, M12 nuts&Many\\ 
    \hline
    Washers&Many\\ 
    \hline
    Aluminum rectangular tube&Several meters (various sizes)\\ 
    \hline
    Aluminum sheets&Many of various sizes and thicknesses(3mm, 5mm and 10mm)\\ 
    \hline
    Carbon fire tube &Several meters\\
    \hline
    PLA (for 3D printers) &About 2 kgs\\ 
    \hline
    Rubber tracks&2\\
    \hline
    Radial bearings&Many (of various types)\\ 
    \hline
    Linear motors (for leg)&2\\
    \hline
    DC motor(for platform)s&2\\
    \hline
    Stepper motors with planetary gearbox&12 (Nema 17 - for arms) + 2 (Nema 11 - for head)\\
    \hline
    Servo motors (for grippers)&2\\
    \hline
    \end{tabular}
    \caption{A short list of materials needed for building Jenny 5. Some lengths and quantities are approximate. Details are given in the assembly manual.}
    \label{tab:materials}
\end{table}

\subsection{Tools required to build Jenny 5}
\label{tools}

Tools required to build Jenny 5 are given in Table \ref{tab:tools_mechanical}.

\begin{table}[ht]
    \centering
    \begin{tabular}{p{2cm}p{3cm}p{6cm}}
    \hline
    Tool&Use&Comment\\
    \hline\hline
Digital caliper&For measuring various sizes&A 150mm and a 300mm caliper are required.\\
    \hline
Mitre Saw&Cutting aluminum profiles&I have a \textit{Bosch GCM 10 MX}, but any other models should work. Make sure that it has a blade which cuts aluminum!\\
    \hline
Column drilling machine&Drilling aluminum, wood parts&I have a \textit{Optimum B16 model}, but any other models should work.\\
    \hline
Hole saw&Cutting holes in aluminum profiles&15-44 mm diameter. I have a \textit{Bosch Progressor} with changing heads.\\ 
    \hline
Screw drivers&screwing...&Various sizes and types (slot, Philips, hex, hex socket) are needed.\\
    \hline
Drill bits for metal&Making various holes in aluminum&2.5 mm, 3.3 mm, 3.9 mm, 6 mm, 8 mm, 12mm diameter.\\ 
    \hline
Wrenches&for tightening screws&Various sizes are needed (mostly for M3, M4, M6, M8 and M12 nuts).\\
    \hline
    \end{tabular}
    \caption{Tools to make the mechanical part of Jenny 5.}
    \label{tab:tools_mechanical}
\end{table}

\subsection{Mobile platform}
\label{mobile_platform_section}

The platform is driven by 2 DC motors with planetary gearbox controlled by a 15 Amps RoboClaw board \cite{ionmc_website}. 

The platform has two rubber tracks.

Several components of the robot are placed on platform: batteries, motor controllers etc.

\begin{figure} 
  \includegraphics[width=\textwidth]{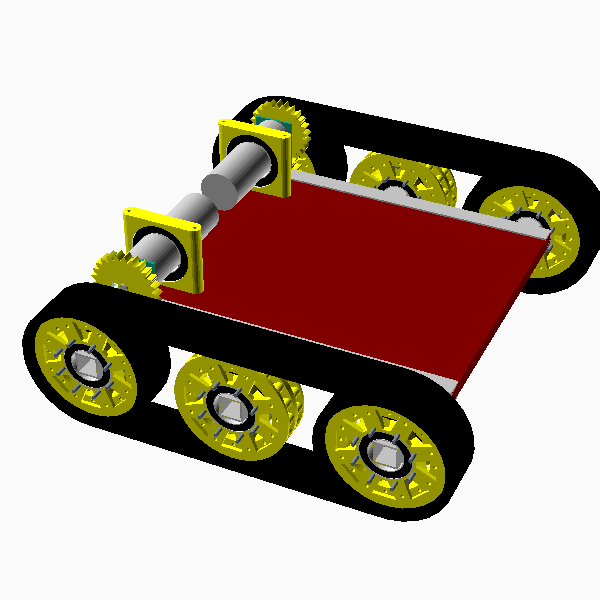}
  \caption{Jenny 5 platform with rubber tracks}
  \label{jenny5_platform_picture}
\end{figure}

The platform is depicted in Figure \ref{jenny5_platform_picture}.

Currently, the platform is quite small and the robot can become unstable (can fall) if abruptly moved when the leg is fully extended. However, if the leg is compressed, the robot is much more stable. So, the idea is to make only moves with low acceleration when the leg is extended. In the near future we plan to add the possibility to move the body back and forth so that we can control the center of gravity easily.

\subsection{Leg}
\label{leg_section}

The leg is driven by two, 750N, linear motors with 100mm stroke, controlled by a 5 Amps RoboClaw board. Note that the motors were disassembled in order to increase the useful stroke.

\begin{figure} 
  \includegraphics[width=\textwidth]{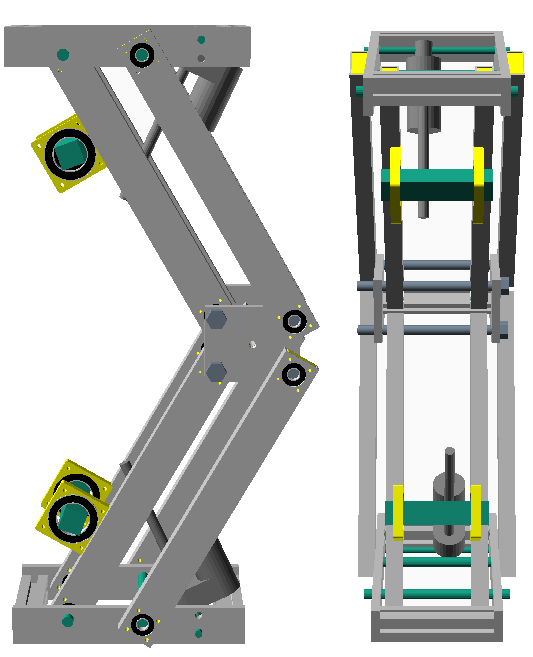}
  \caption{Jenny 5 leg. Lateral and frontal view.}
  \label{leg_design}
\end{figure}

In Figure \ref{leg_design} we have a picture with the lateral and frontal leg design view.

The leg can be compressed so that arms can grab things from ground.

\subsection{Arms}
\label{arms_section}

Each arm has 6 stepper motors with various gearbox reductions (27:1 and 50:1). Joint position is read using a magnetic rotation sensor (AS 5147 \cite{as5147}). Motors and sensors are controlled with a Pololu A-Star board \cite{pololu_a_star}.

The 50:1 motors are used for lifting the entire arm and for moving the elbow. Theoretically they are capable of a lot of force. According to the specifications \cite{stepperonline} the motors (without gearbox) have 52N.cm torque. That multiplied with 50:1 reduction means 2600N.cm. The efficiency of gearboxes is 73\%, so the useful torque is around 1900N.cm. The motor does not rotate the arm directly, but through the help of a 47:14 pulley reduction and this means that the final torque is around 6370 N.cm. At a distance of 70 cm (arm length) it means a 91 N.cm, which means that it can lift about 9kgs at 70cm. Note that this is a theoretical upper bound because the gearbox does not support more than 600N.cm before damage can occur \cite{stepperonline}, so the practical upper bound is below 3kg at 70 cm. Also, note that neither friction nor arm's weight were taken into account here. However the arm weight can be canceled if a spring is used. One more aspect to consider is that the torque of stepper motor decreases at high speeds \cite{torque_vs_speed}, so again the practical upper bound for torque is lower than the one computed before.

Still, in our experiments (see for instance movie at \url{https://www.youtube.com/watch?v=Rc-ppA9-12I} we have been able to lift more than 3.5kg at 40cm. This means a torque of at least 1400N.cm and no gearbox damage was observed in long term.

Another problem is that 50:1 gearboxes are difficult to back-drive. In the near future we plan to replace the 50:1 gearboxes with 27:1 ones because the maximum allowed torque before damage can occur is similar.

Regarding the speed of the 50:1 gearbox motors we have observed and made the following computations: The maximum speed that we were able to obtain with the motor (no gearboxes attached was 1500 steps/second. After that speed the motor just stops). Knowing that the motor has 1.8 degrees steps, it means that we can about 7.5 complete rotations in 1 second. After the 50:1 gearbox we can make 0.15 of a complete rotation in 1 second. After the external 47:14 reduction we make 0.044 of a complete rotation in 1 second, that is about 16 degrees in 1 second. Thus, a complete up-down move (180 degrees) of the arm will take about 11 seconds. We are currently investigating other stepper motor drivers to see if more steps per second are possible.

Each arm is controlled independently.
The electronic boards are placed on each arm separately.

\begin{figure} 
  \includegraphics[width=\textwidth]{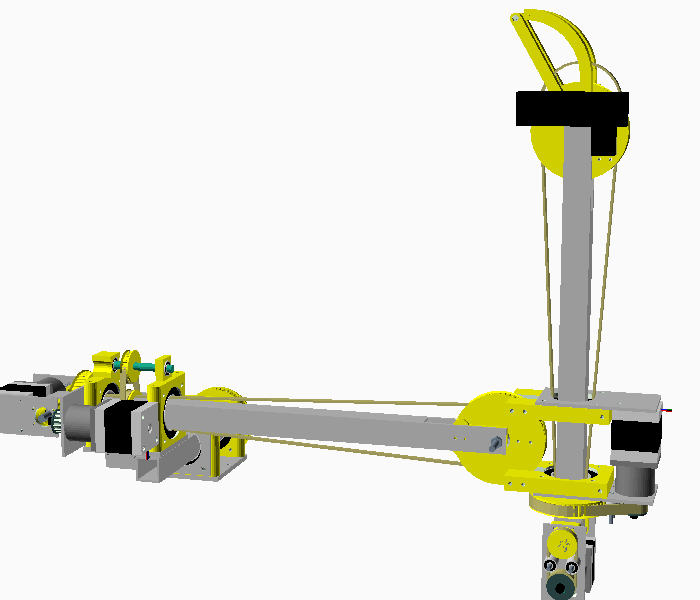}
  \caption{Jenny 5 arm.}
  \label{arm_image}
\end{figure}

The arm is depicted in Figure \ref{arm_image}.

\subsection{Gripper}
\label{gripper}

Each arm has attached a gripper which is moved by a servo motor. The gripper is connected to the same board as the rest of the arm. The gripper has attached a web-cam used for recognizing the objects in the front of it.

\begin{figure} 
  \includegraphics[width=\textwidth]{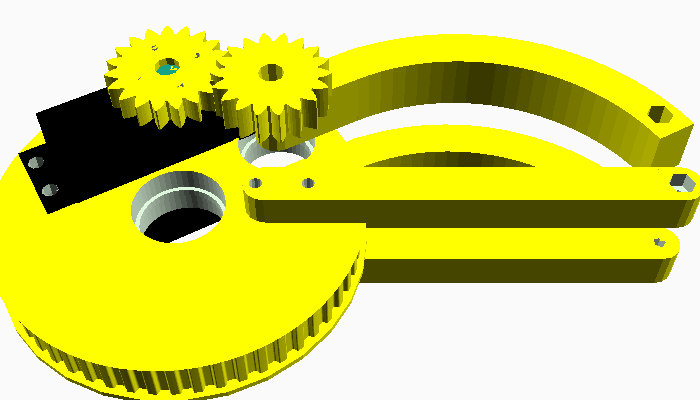}
  \caption{Jenny 5 gripper.}
  \label{gripper_image}
\end{figure}

The gripper is depicted in Figure \ref{gripper_image}.

\subsection{Body}
\label{body_section}

The body is a carbon-fiber frame which has support for 2 arms and connectors for leg (at the bottom) and head (at the top).

\begin{figure} 
  \includegraphics[width=\textwidth]{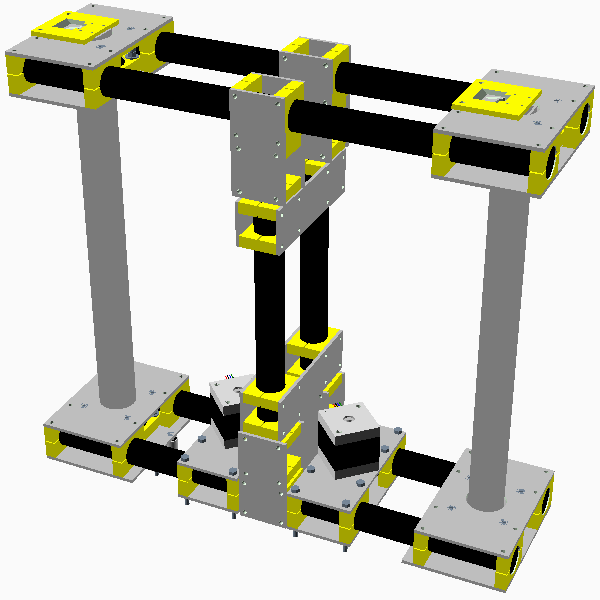}
  \caption{Jenny 5 body.}
  \label{body_image}
\end{figure}

Body is depicted in Figure \ref{body_image}.

\subsection{Head}
\label{head_section}

The head has 2 degrees of freedom ensured by two Nema 11 stepper motors. Position of the motors is read by 2 AS5147 sensors.

The head has a web-cam for object detection and an ultrasonic sensor for distance measurement. 

All head components, except the camera, are connected to an Arduino Nano board.

\newpage
\section{Electronics}
\label{electronics}

This section describes the electronics components utilized by Jenny 5. First we list the materials and tools required to build the electrical part of the robot. The we describe the custom PCBs build for arms and head.

\subsection{Electrical / electronic materials}

A list of materials is given in Table \ref{tab:electrical_materials}. Note that motors were listed in Table \ref{tab:materials} (see the hardware section \ref{hardware}).

\begin{table}[ht]
    \centering
    \begin{tabular}{p{3cm} p{1cm} p{6.5cm}}
    \hline
    Component&Quantity&Comment\\
    \hline\hline
         Arduino Nano&1&for head\\
         \hline
         RoboClaw 7Amps&1&Controls the leg motors\\
         \hline
         RoboClaw 15Amps&1&Controls the base platform motors\\
         \hline
         Pololu A-Star mini&2&Controls the arms. We have chosen this board instead of Arduino Nano because A-Star has more pins.\\
         \hline
         Custom PCBs&4&For arms and head. More custom PCBs for platform and legs are under development.\\
         \hline
         Pololu A4988 stepper driver&15& for driving motors on arms and head.\\
    \hline
    AS5147 rotary encoder \cite{as5147}&14&for reading the position of each arm and head articulation\\
    \hline
    Logitech C920 webcam&3&for video capture. It is place on head and arms (for detecting the grabbed objects).\\
    \hline
    HC-SR04 ultrasonic&1&for raw obstacles detection\\
    \hline
         wires&Many&AWG 18 (for leg and platform), AWG 22 (for motors and sensors on arms and head).\\
         \hline
         resistors&Several&\\
         \hline
         capacitors&Many&\\
         \hline
         connectors&Many&\\
    \hline 
    \end{tabular}
    \caption{Electrical materials required to build Jenny 5.}
    \label{tab:electrical_materials}
\end{table}

This section describes the electronic used to control the robot. 

\subsection{Tools}
\label{electronic_tools}

Tools requires to build the electrical parts for robot are given in Table \ref{tab:tools_electrical}.

\begin{table}[ht]
    \centering
    \begin{tabular}{p{2.8cm}p{5.5cm}p{3cm}}
    \hline
Tool&Use&Comment\\
    \hline\hline
Digital multimeter&for measuring currents and voltages& \\
    \hline
Soldering iron&for soldering electrical components&A 20-30W iron is OK.\\
\hline
    \end{tabular}
    \caption{Tools required to build the electronics / electrical part of Jenny 5}
    \label{tab:tools_electrical}
\end{table}

\subsection{Arm custom PCB}

\begin{figure} 
  \includegraphics[height=0.9\textheight]{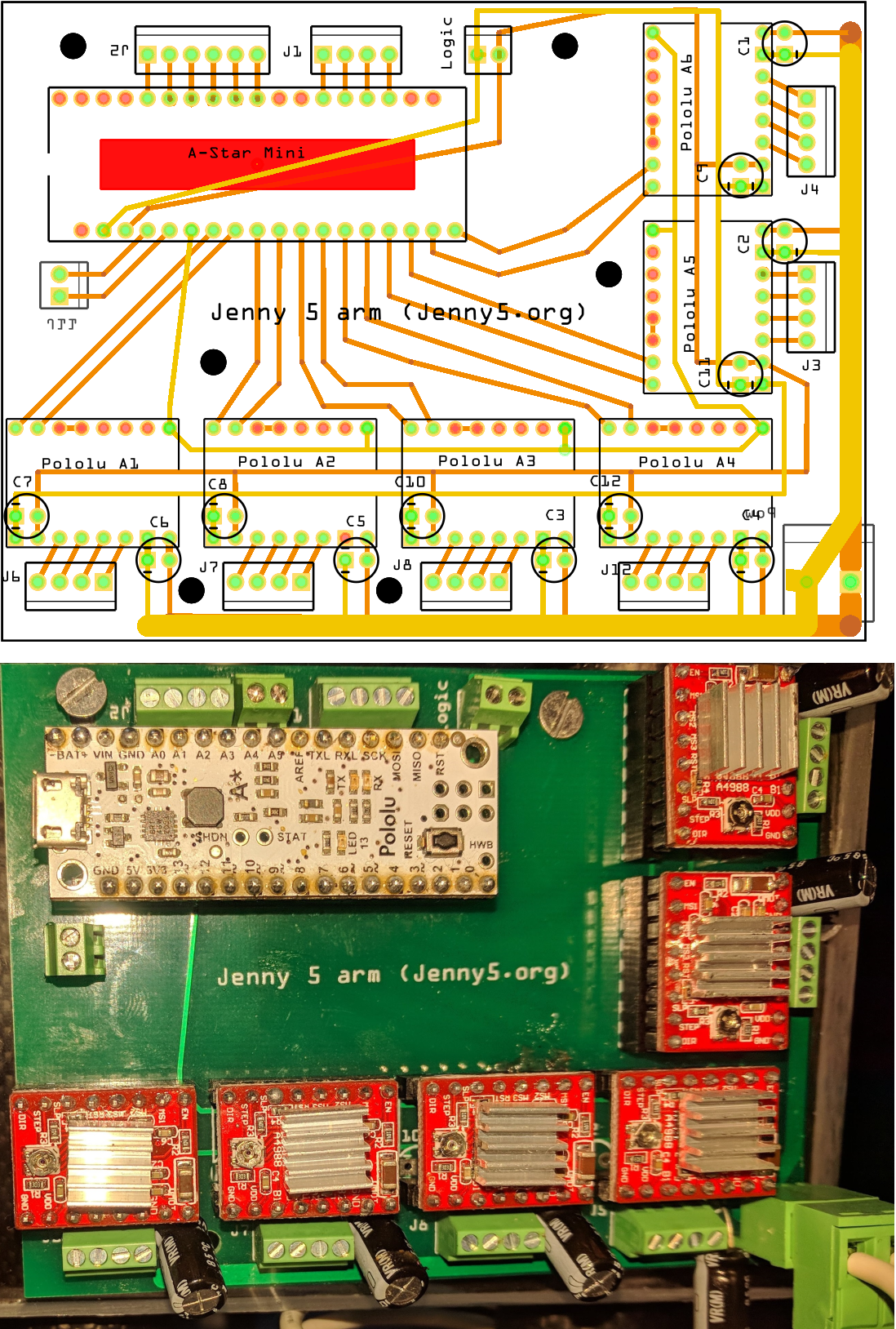}
  \caption{Jenny 5 arm electronics. Design view (top) and actual implementation (bottom).}
  \label{arm_electronics}
\end{figure}

The PCB for arm is depicted in Figure \ref{arm_electronics}. The electronics was designed using Fritzing \cite{fritzing}.

\newpage
\section{The brain}
\label{brain}

The brain of the robot is a 13” laptop, with i7-6700 processor, 128GB SSD, and 4GB RAM and is placed on the platform. The laptop has 3 USB-A ports (more ports is better, otherwise some USB-hubs are needed).

The laptop sends commands to A-star, Arduino and RoboClaw boards and read images from webcams. 

Obviously, any other laptop can be utilized, preferably small and powerful. A low-cost computer like Raspberry Pi \cite{raspberry_pi} can be used too if the robot is not doing any high costs tasks like computer vision.

\section{Power supply}
\label{power_supply}

The Jenny 5 robot is powered by multiple Li-Po rechargeable batteries. Currently a 16 Ah battery powers both arms and the head and a 10 Ah battery powers the platform and the leg. The laptop is powered by its own battery.

\newpage
\section{The software controlling the robot}
\label{jenny5_software}

The robot components are controlled by various programs. Arms and head run the \textit{Scufy} - the Arduino firmware which accepts commands sent from PC with the help of  \textit{Scufy Lib}. The RoboClaw boards run a proprietary firmware and accepts commands sent by a C++ library running on a PC.

\subsection{Scufy - the Arduino firmware}
\label{arduino_firmware}

A-star and Arduino boards run a specially crafted firmware called \textit{Scufy} (see the source code in reference \cite{scufy_source_code}) which is able to control multiple motors and read a variety of sensors: buttons, AS5147 magnetic rotary encoders, HC-SR04 ultrasonic, potentiometers, infrared, TeraRanger One etc. 

The firmware is asynchronous and does not block the board when performing long running operations (such as moving a motor to a new position (which can take sometimes several seconds) or reading an ultrasonic sensor).

\textit{Scufy} firmware accepts commands sent through the serial port.

All commands are terminated by \# character.

Almost all commands send a response to the serial port. The response is terminated by \# character.

Commands related to motors and sensors required an object (controller) to be created internally for each type of motor or sensor. This controller stores how many sensors of that type we have and more information about the Arduino pins where that motor/sensor is connected. etc. So, before moving motors or reading sensors the user must create a controller for them. The motors / sensors inside a controller are indexed from 0.

The list of most important \textit{Scufy} commands is given below. The user should read the entire / updated list of commands from the source code of the program.

\subsubsection{General commands}

\begin{itemize}
    \item 
	$T$\#
	- Test connection. 
	- Outputs $T$\#.
	
    \item 
	$V$\#
	- Outputs version string (year.month.day.build\_number). eg: $V2019.05.10.0$\#.
\end{itemize}

\subsubsection{Create controllers commands}

\begin{itemize}
    \item 
	$CS\ n\ dir_0\ step_0\ en_0\ dir_1\ step_1\ en_1\ ...\ dir_{n-1} step_{n-1}\ en_{n-1}\#$
	
	Creates the stepper motors controller and set some of its parameters.
	
	$n$ is the number of motors, $dir_k$, $step_k$, $en_k$ are the Arduino pins which command the direction, step and the enable state.
	
	Outputs $CS$\# when done.
	
	Example: $CS\ 3\ 5\ 4\ 12\ 7\ 6\ 12\ 9\ 8\ 12$\#

	\item $CV\ n\ pin_0\ pin_1\ ...\ pin_{n-1}$\# 
	
	Creates the servo motors controller and set its pins. 
	
	$n$ is the number of motors, $pin_k$ are Arduino pins where the motors are connected.
	
	Outputs $CV$\# when done.

	\item $CA\ n\ pin_0\ pin_1\ ...\ p_{n-1}$\#
	
	Creates the AS5147 controller. 
	
	$n$ is the number of sensors, $p_k$ are Arduino pins where the sensors are connected.
	
	Outputs $CA$\# when done.
	
	Example: $CA\ 3\ 18\ 19\ 20$\#

\end{itemize}

\subsubsection{Attach sensors to motors}

\begin{itemize}

	\item $ASx\ n\ P_y\ end_1\ end_2\ home\ direction\ A_k\ end_1\ end_2\ home\ direction$\# 
	
	Attach, to stepper motor $x$, a list of n sensors (like Potentiometer y, Button z, AS5147 etc).
	
	$y$ is the sensor index in the list of sensors of that type.
	
	$end_1$ and $end_2$ specify the sensor angular position guarding the motor movement.
	
	$home$ specifies the home position of the motor.
	
	$direction$ specifies if the increasing values for motor will also increase the values of the sensor.
	
	Outputs $ASx$\# when done.
	
	Example: $AS0\ 1\ A0\ 280\ 320\ 300\ 1$\#
	
\end{itemize}

\subsubsection{Stepper Commands}

\begin{itemize}
	\item $SMx\ y$\#
	
	Moves stepper motor $x$ with $y$ steps. If $y$ is negative the motor runs in the opposite direction. The motor remains locked at the end of the movement.
	
	The first motor has index 0.
	
	Outputs $SMx\ d$\# when motor rotation is over. If the movement was complete, then $d$ is 0, otherwise is the distance to go.
	
	Example: $SM1\ 100$\#

	\item $SHx$\#
	
	Moves stepper motor $x$ to the home position. 
	
	The first sensor in the list of sensors will establish the home position. The motor does nothing if no sensor is attached. 
	
	Outputs $SHx$\# when done.

	\item $SDx$\#
	
	Disables stepper motor $x$. 
	
	Outputs $SDx$\# when done.

	\item $SLx$\#
	
	Locks stepper motor $x$ in the current position. 
	
	Outputs $SLx$\# when done.

	\item $SSx\ speed\ acceleration$\#
	
	Sets the speed of stepper motor $x$ to $speed$ and the acceleration to $acceleration$.
	
	Outputs $SSx$\# when done.

	\item $STx$\#
	
	Stops motor $x$.
	
	Outputs $STx$\# when done.

	\item $SGx\ position$\#
	
	Moves stepper motor $x$ to $position$ sensor position. The first sensor in list will give the position.
	
	Outputs $SMx\ d$\# when motor rotation is over. If the movement was complete, then $d$ is 0, otherwise is the distance left to go.
	
	Example: $SG1\ 100$\#

\end{itemize}

\subsubsection{Servo commands}

\begin{itemize}
	\item $VMx\ position$\# 
	
	Moves servo motor $x$ to $position$.
	
	Outputs $VMx\ d$\# when done. If the move is completed $d$ is 0, otherwise $d$ is 1.

	\item $VHx$\#
	
	Moves servo motor $x$ to the home position. Home position is given by the first sensor in the list of attached sensors.  
	Outputs $VHx$\#.
	
\end{itemize}

\subsubsection{Read sensors commands}

\begin{itemize}
	\item $RAx$\#
	
	Read the value of AS5147 sensor. 
	
	Outputs $RAx\ angle$\#.

\end{itemize}

\subsubsection{Others}
	
	\begin{itemize}
	\item $E$\#
	
	The firmware can output this string if there is something wrong with a given command.

	\item $I information$\#
	
	The firmware can output this string containing useful information about the progress of a command.

\end{itemize}

\subsection{Scufy Lib - the PC library}
\label{Arduino_control_library}

\textit{Scufy Lib} \cite{jenny5_arduino_control_library} is a C++ library that sends commands (as strings) and reads results to / from the electronic boards powering the arms, head etc. Communication between on board computers and the electronic boards happens on a serial port.

The following methods are a part of the \textit{Scufy Lib}. The user should read the entire list of commands from the source code of the library.

\subsubsection{General commands}

    \begin{lstlisting}[breaklines, language = C++]
    
// connects to given serial port
bool connect(const char* port, int baud_rate);

// test if the connection is open; For testing if the Arduino is alive please send_is_alive method and wait for IS_ALIVE_EVENT event
bool is_open(void);
	
// close serial connection
void close_connection(void);
	
// sends (to Arduino) a command (T#) for testing if the connection is alive
// when the Arduino will respond, the event will be added in the list
void send_is_alive(void);

\end{lstlisting}

\subsubsection{Events processing}

    \begin{lstlisting}[breaklines, language = C++]

// reads data from serial and updates the list of received events from Arduino
// this should be called frequently from the main loop of the program in order to read the data received from Arduino
bool update_commands_from_serial(void);

// search in the list of events for a particular event type
// it returns true if the event is found in list
// the first occurrence of the event is removed from the list
bool query_for_event(int event_type);
	
// search in the list of events for a particular event type
// it returns true if the event is found in list
// param1 parameter will be set to the information from the param1 member of the event
bool query_for_event(int event_type, int* param1);

// search in the list of events for a particular event type
// it returns true if the event is found in list
// param1 parameter will be set to the information from the param1 member of the event
// param2 parameter will be set to the information from the param2 member of the event
bool query_for_event(int event_type, int *param1, intptr_t *param2);

// search in the list of events for a particular event type
// returns true if the event type matches and if the param1 member is equal to the parameter given to this function
bool query_for_event(int event_type, int param1);

\end{lstlisting}

\subsubsection{Create commands}
	
\begin{lstlisting}[breaklines, language = C++]

// sends (to Arduino) a command for creating a stepper motor controller
// several arrays of pin indexes for direction, step and enable must be specified
// this method should be called once at the beginning of the program
// calling it multiple times is allowed, but this will only fragment the Arduino memory
void send_create_stepper_motors(int num_motors, int* dir_pins, int* step_pins, int* enable_pins);
	
// sends a command for creating a AS5147s controller
// this method should be called once at the beginning of the program
// calling it multiple times is allowed, but this will only fragment the Arduino memory
void send_create_as5147s(int num_as5147s, int* out_pins);

// sends a command for creating a Tera Ranger One controller
// this method should be called once
// only one sensor is permitted per Arduino board
void send_create_tera_ranger_one(void);


\end{lstlisting}

\subsubsection{Stepper motor movements}

\begin{lstlisting}[breaklines, language = C++]

// sends (to Arduino) a command for moving a stepper motor to home position
void send_go_home_stepper_motor(int motor_index);

// sends a command for moving a motor with a given number of steps
void send_move_stepper_motor(int motor_index, int num_steps);
	
// sends a command for moving two motors
void send_move_stepper_motor2(int motor_index1, int num_steps1, int motor_index2, int num_steps2);
	
// sends a command for moving three motors
void send_move_stepper_motor3(int motor_index1, int num_steps1, int motor_index2, int num_steps2, int motor_index3, int num_steps3);
	
// sends a command for moving four motors
void send_move_stepper_motor4(int motor_index1, int num_steps1, int motor_index2, int num_steps2, int motor_index3, int num_steps3, int motor_index4, int num_steps4);
	
// sends a command for moving multiple motors
void send_move_stepper_motor_array(int num_motors, int* motor_index, int *num_steps);

// sends a command for stopping a stepper motor
void send_stop_stepper_motor(int motor_index);

// sends a command for moving a motor to a new sensor position
void send_stepper_motor_goto_sensor_position(int motor_index, int sensor_position);

// sends a command for blocking a motor to current position
void send_lock_stepper_motor(int motor_index);
	
// sends a command for disabling a motor
void send_disable_stepper_motor(int motor_index);

// sends a command for setting the speed and acceleration of a given motor
void send_set_stepper_motor_speed_and_acceleration(int motor_index, int motor_speed, int motor_acceleration);

\end{lstlisting}

\subsubsection{Attach sensors to motors}
	
\begin{lstlisting}[breaklines, language = C++]

// sends (to Arduino) a command for attaching several sensors to a given motor
void send_attach_sensors_to_stepper_motor(int motor_index, 
    int num_potentiometers, int *potentiometers_index, 
	int* _low, int* _high, int *home, int *_direction, 
	int num_AS5147s, int *AS5147_index,
	int* AS5147_low, int* AS5147_high, int *AS5147_home, int *AS5147_direction,
	int num_infrared, int *infrared_index,
	int num_buttons, int *buttons_index, int *button_direction
);

// sends a command for reading removing all attached sensors of a motor
void send_remove_attached_sensors_from_stepper_motor(int motor_index);
\end{lstlisting}

\subsubsection{Reading Sensors}
	
\begin{lstlisting}[breaklines, language = C++]

// sends (to Arduino) a command for reading a AS5147 position
void send_get_AS5147_position(int sensor_index);


	\end{lstlisting}

\subsubsection{State}
	
\begin{lstlisting}[breaklines, language = C++]

// returns the state of a motor
int get_stepper_motor_state(int motor_index);

// sets the state of a motor
void set_stepper_motor_state(int motor_index, int new_state);

\end{lstlisting}

\subsubsection{Scufy Lib events}

Strings received from \textit{Scufy} firmware are translated by \textit{update\_commands\_from\_serial()} to events which are stored into a queue. Each event has a particular type. A short list of event types is given in Table \ref{tab:messages}. For more events the user is encouraged to read the \textit{scufy\_events.h} file from the \textit{Scufy Lib} repository.

\begin{table}[ht]
    \centering
    \begin{tabular}{p{4cm} p{7cm}}
         Event&Meaning  \\
         \hline\hline
         IS\_ALIVE\_EVENT&Received if the \textit{Scufy} firmware responded to a $T\#$ command\\
         \hline   
         STEPPER\_MOTORS\_ CONTROLLER\_CREATED\_EVENT&Received after the stepper motor controller has been created.\\
         \hline
         ATTACH\_SENSORS\_ EVENT&Received after sensors have been attached to motors\\
    \hline
    AS5147\_READ\_EVENT&Received after the sensor has been successfully read.\\
         \hline
         STEPPER\_MOTOR\_MOVE\_ DONE\_EVENT&Received after the motor has finished the requested move.\\
         \hline
    \end{tabular}
    \caption{Some of the events of \textit{Scufy Lib}. }
    \label{tab:messages}
\end{table}

\subsubsection{Example of utilization}

In this section we give some example of utilization for the Scufy Lib.

After sending a command to Arduino firmware, the PC program should wait for an answer in an asynchronous way. Since the answer is not instantaneous the programmer should create a loop where it waits for an answer. The basic idea is the following:

\begin{itemize}
\item
send a command,

\item
use \textit{update\_commands\_from\_serial()} to extract strings sent by firmware,

\item use \textit{query\_for\_event()} to determine if a particular event has been received.
\end{itemize}

Below is an example of code which creates a stepper controller. This code is actually used by the server to create a controller for the left arm of the robot \cite{jenny5_arduino_control_library}.

\begin{lstlisting}[breaklines, language = C++]
bool t_left_arm_controller::create_stepper_motors_controller(char* error_string)
{
	int left_arm_motors_dir_pins[6] = { 5, 7, 9, 11, 3, 1 };
	int left_arm_motors_step_pins[6] = { 4, 6, 8, 10, 2, 0 };
	int left_arm_motors_enable_pins[6] = { 12, 12, 12, 12, 12, 12 };
	arduino_controller.send_create_stepper_motors(6, left_arm_motors_dir_pins, left_arm_motors_step_pins, left_arm_motors_enable_pins);

	bool motors_controller_created = false;
	clock_t start_time = clock();
	while (1) {
		if (!arduino_controller.update_commands_from_serial())
			Sleep(5); // no new data from serial ... we make a little pause so that we don't kill the processor

		if (arduino_controller.query_for_event(STEPPER_MOTORS_CONTROLLER_CREATED_EVENT, 0)) {  // have we received the event from Serial ?
			motors_controller_created = true;
			break;
		}

		// measure the passed time 
		clock_t end_time = clock();

		double wait_time = (double)(end_time - start_time) / CLOCKS_PER_SEC;
		// if more than 3 seconds then game over
		if (wait_time > NUM_SECONDS_TO_WAIT_FOR_CONNECTION) {
			if (!motors_controller_created)
				sprintf(error_string, "Cannot create left arm's motors controller! Game over!\n");
			return false;
		}
	}
	return true;
}
//----------------------------------------------------------------

\end{lstlisting}

\subsection{RoboClaw control library}
\label{Roboclaw_library}

RoboClaw library \cite{roboclaw_control_library_code} is C++ library which send commands, on a serial port, to the RoboClaw board. 

Currently the platform and the leg are controlled by RoboClaw boards. 

The following functions are included in library:

\begin{lstlisting}[language = c++, breaklines]
	// returns the library version
	// the caller must not delete the pointer
	const char* get_library_version(void);

	bool connect(const char* port, int baud_rate);
	void close_connection(void);
	bool is_open(void);

	// Read the board temperature. //Value returned is in 10ths of degrees.
	double get_board_temperature(void);

	// Read the main battery voltage level connected to B+ and B- terminals
	double get_main_battery_voltage(void);

	// Read RoboClaw firmware version. 
	// Returns up to 48 bytes
	// (depending on the RoboClaw model) and 
	// is terminated by a line feed character and a null character.
	void get_firmware_version(char *firmware_version);

	// Drive motor 1 forward.
	//Valid data range is 0 - 127. 
	// A value of 127 = full speed forward, 
	// 64 = about half speed forward and 0 = full stop.
	bool drive_forward_M1(unsigned char speed);
	// Drive motor 2 forward.
	// Valid data range is 0 - 127. 
	// A value of 127 = full speed forward, 
	// 64 = about half speed forward and 0 = full stop.
	bool drive_forward_M2(unsigned char speed);
	bool drive_backward_M1(unsigned char speed);
	bool drive_backward_M2(unsigned char speed);

	// Read the current draw from each motor in 10ma increments.
	// The amps value is calculated by	dividing the value by 100.
	void get_motors_current_consumption(double &current_motor_1, double &current_motor_2);

	// Read the current PWM output values for the motor channels.
	// The values returned are + -32767.
	// The duty cycle percent is calculated by dividing the value by 327.67.
	void read_motor_PWM(double &pwm_motor_1, double &pwm_motor_2);

	// The duty value is signed and the range is:
	// - 32768 to + 32767(eg. + -100 % duty).
	// The acceleration value range is 0 to 655359 
	// (eg. maximum acceleration rate is - 100 % to 100 % in 100ms).
	bool drive_M1_with_signed_duty_and_acceleration(int16_t duty, uint32_t acceleration);

	// The duty value is signed and the range is:
	// - 32768 to + 32767 (eg. + -100 % duty).
	// The acceleration value range is 0 to 655359 
	// (eg. maximum acceleration rate is - 100 % to 100 % in 100ms).
	bool drive_M2_with_signed_duty_and_acceleration(int16_t duty, uint32_t acceleration);

	// Set Motor 1 Maximum Current Limit. 
	//Current value is in 10ma units.
	//  To calculate multiply current limit by 100.
	bool set_M1_max_current_limit(double c_max);

	// Set Motor 1 Maximum Current Limit. 
	// Current value is in 10ma units.
	//  To calculate multiply current limit by 100.
	bool set_M2_max_current_limit(double c_max);

\end{lstlisting}

\subsection{HTML 5 client and PC WebSocket server}
\label{Mobile_client_and_PC_server}

The robot can be manually controlled by an \textit{HTML5} application running within the browser of a smartphone. The HTML5 application connects to the server running on the robot. The server is the one that actually execute the commands (move motors, read sensors) etc.

The server is built on a top of a light WebSocket server (single source file) written by Eduard \cb{S}uică \cite{eduard_github}. The server uses \textit{TLSe} library \cite{tlse} for the secured communication protocol.

The server requires a certificate to run. A sample certificate has been generated and stored in the \textit{certificates} folder of the server. This will work with no problems on smartphones running Android. However, for iOS a new certificate must be generated. More details on how to do this for iOS can be found in reference \cite{ios_certificate}.

The Jenny 5 web client allows to control one motor (as in the case of arms) or maximum two motors (as in the case of platform, leg and head) at a time. In such scenario the utilization of the application is very simple: 

\begin{itemize}
\item
the user presses a button (to select the motor that he wants to move),
\item
the tilts the smartphone, 
\item
the client application read the gyroscope of the smartphone,
\item
the client application send the angle to the server on the robot,
\item
and the robot acts accordingly. 
\end{itemize}

The client can request a picture from robot and then displays it in the browser window.

The web client also accepts voice commands. This feature is implemented by using \textit{Speech Recognition} from HTML 5 \cite{speech_recognition_html5}. 

\begin{figure} 
  \includegraphics[height=0.9\textheight]{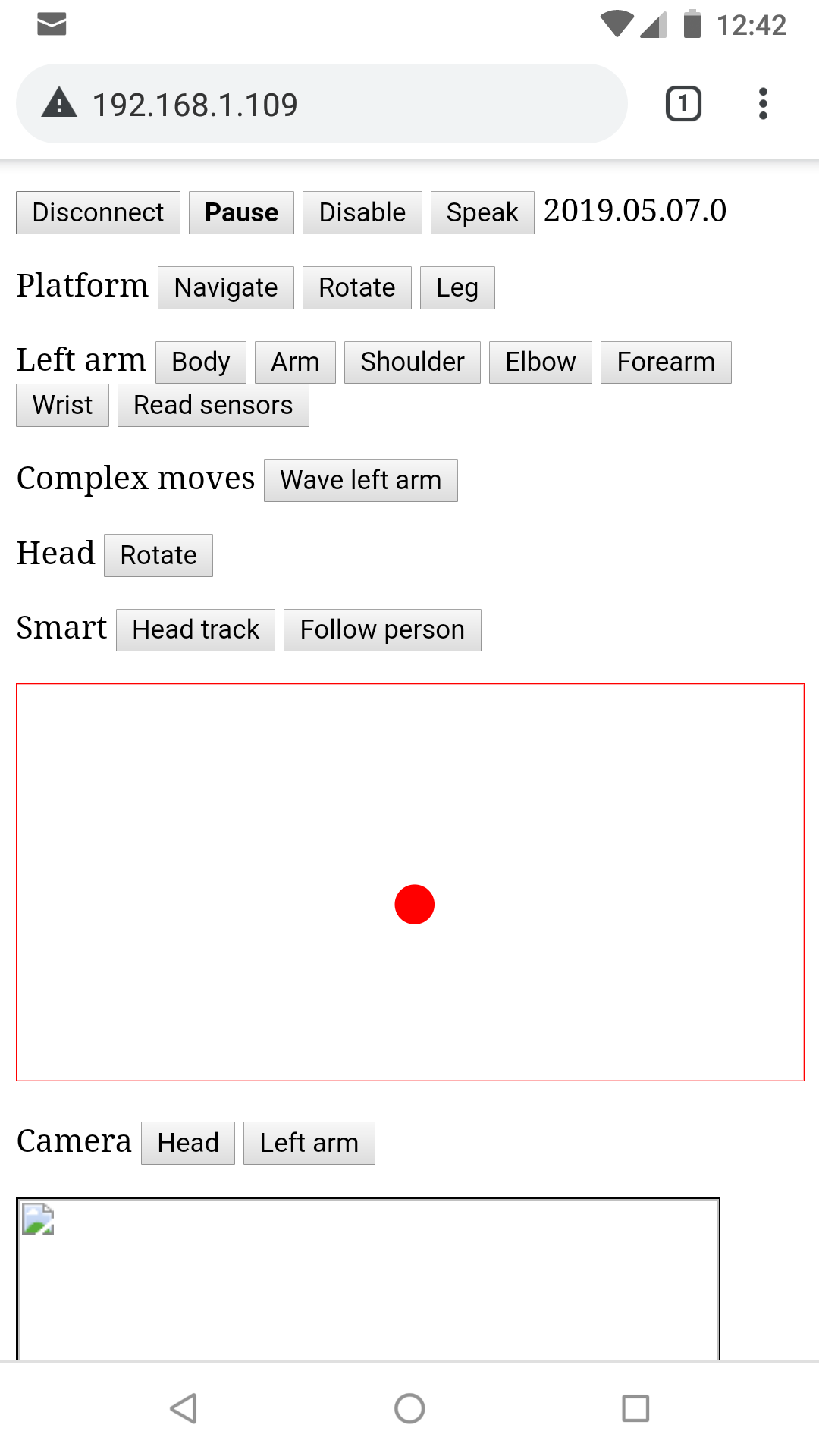}
  \caption{Jenny 5 HTML 5 web client. Red circle indicates the tilt of the smartphone.}
  \label{html5_client}
\end{figure}

An screenshot of the HTML5 client is depicted in Figure \ref{html5_client}.

\subsection{Intelligent algorithms}
\label{Intelligent_algorithms}

Currently there are two intelligent algorithms implemented on the robot: 

\begin{itemize}
    \item 
An algorithm which finds the closest face and moves the head’s motors in order to center it on the camera view. Only the head is involved in this operation.

    \item 
An algorithm which follows the closest person. First it detects the closest face and then if the person is too close, the robot moves backward, otherwise it moves forward.
\end{itemize}

Both algorithm uses the OpenCV \cite{opencv_website} for face detection.

\newpage
\section{Weaknesses and future development}
\label{Future_work}

Jenny 5 robot is under active development. 

During the development of the Jenny 5 robot we have observed several weaknesses that we plan to address in future iterations.

Some short term ideas are listed in Table \ref{tab:future_work}.

\begin{table}[htbp]
\caption{Future work on Jenny 5.}
\label{tab:future_work}
\begin{tabular}
{p{120pt}p{200pt}}

\hline
Task & Purpose\\
  \hline\hline

Rotatable body&Could reach farther objects without moving the base platform\\
\hline
Springs on upper arm-body articulation&Will lift heavier weights\\
\hline
Two legs&More stability and could climb stairs\\
\hline

Intelligent algorithm for grabbing a bottle&More intelligence\\
\hline
Change speed for stepper motors during run&Finer control of the arms\\
\hline
LiDAR on platform and head&Better detection of obstacles\\
\hline
Visualizing the LiDAR on the client application&Can send the robot far away and see what it sees\\
\hline
Visualizing the camera (real time video) on the client application&Can send the robot far away and see what it sees\\
\hline

Arms electronics build around Arduino Mega&Could attach more sensors to arms\\
\hline

Custom made PCBs for leg and platform&Can have own software and can attach more sensors to them\\
\hline
Carbon fiber sheets&To reduce the total weight of the robot\\
\hline
Body movable back and forward&To make the robot more stable on inclined plane\\
\hline
Magnetic rotary sensors on leg&To determine the exact position of leg\\
\hline
Magnetic rotary sensors on platform gears&To determine the true speed of the platform\\
\hline
Gyroscope on platform&To determine if the robot will fall\\ 
\hline

\end{tabular}
\end{table}

\newpage
\section*{Acknowledgement}

The author likes to thank to the following of his students which have helped him while working to older prototypes of Jenny 5:

\begin{itemize}
    \item 
2017 students from Intelligent Robots class: Alexandru Donea, Daniela Oni\c ta, Flaviu Suciu, Tudor Samuila, Daniel Leah, Leontin Neam\c tu, Todor U\c t, Marius Penciu, Florin Jurj, Alin Simina, Mihaela Ro\c sca.

    \item 
2016 students from Intelligent Robots class: Pop Ioana, Mure\c san Andreea, Kisari Andrei, Turtoi Cristian, Mate\c s Ciprian, Chereghi Adrian, Buciuman Mircea, Bochi\c s Bogdan.

    \item 
2015 students from Intelligent Robots class:  Andra Ristei, Horea Mure\c san, Baciu Iulian, Bica Ioana, Bonte Aurelian, Ilie\c s Daniel, Lonhard Cristian, Lungana Niculescu Alexandru Mihai, Marian Cristian, Sorban Timea, Tiperciuc Corvin.
\end{itemize}

\end{document}